%% file: main.tex
\documentclass[10pt,twocolumn,letterpaper]{article}

\usepackage[pagenumbers]{cvpr} 

\usepackage{graphicx}   
\usepackage{tabularx}   
\usepackage{booktabs}   
\usepackage{pifont}
\usepackage{booktabs}
\usepackage{multirow}
\usepackage{array}
\usepackage{cuted} 
\usepackage{caption}

\usepackage{colortbl}
\usepackage{xcolor}
\usepackage{booktabs}
\definecolor{first}{RGB}{255, 255, 153}
\definecolor{second}{RGB}{200, 200, 200}
\usepackage{makecell} 
\usepackage{dsfont}
\usepackage{bbm}

\usepackage{xcolor}
\definecolor{rosepink}{RGB}{220,20,147} 
\usepackage[colorlinks=true, urlcolor=rosepink]{hyperref}

\newcommand{\tightpara}[1]{\vspace{4pt}\noindent\textbf{#1}\,}

\usepackage{tikz}
\usetikzlibrary{shadows.blur}

\usepackage{tcolorbox}
\tcbuselibrary{skins}

\definecolor{uiHeader}{HTML}{2F3B46}

\newtcolorbox{promptbox}[1]{%
  enhanced,               
  colback=white,           
  colframe=black!40,     
  boxrule=1pt,           
  arc=3pt,              
  left=6pt, right=6pt, top=4pt, bottom=4pt,
  title={#1},
  attach boxed title to top left={xshift=0mm,yshift=-2mm},
  boxed title style={
    colback=uiHeader, colframe=uiHeader,
    boxrule=0pt,
    sharp corners,
    top=2pt, bottom=2pt,
    left=6pt, right=6pt,
  },
  fonttitle=\bfseries\large\color{white},
before upper={\normalfont\normalsize\color{black}\setlength{\parskip}{2pt}},
}

\usepackage{listings}

\input{preamble}

\def\paperID{8140} 
\def\confName{CVPR}
\def\confYear{2026}

\title{ViBES: A Conversational Agent with Behaviorally-Intelligent 3D Virtual Body}

\author{
Juze Zhang$^1$ \hspace{1.5mm} Changan Chen$^1$ \hspace{1.5mm} Xin Chen$^2$ \hspace{1.5mm} Heng Yu$^1$\hspace{1.5mm} Tiange Xiang$^1$ \hspace{1.5mm} Ali Sartaz Khan$^1$\hspace{1.5mm} \\ Shrinidhi K. Lakshmikanth$^1$\hspace{1.5mm} Ehsan Adeli$^1$
\\
$^1$Stanford University \hspace{3mm}
$^2$ByteDance
}

\begin{document}
\maketitle

\begin{strip}\centering
    \vspace*{-1.8cm}
    \captionsetup{type=figure}
    \includegraphics[width=1.0\textwidth]{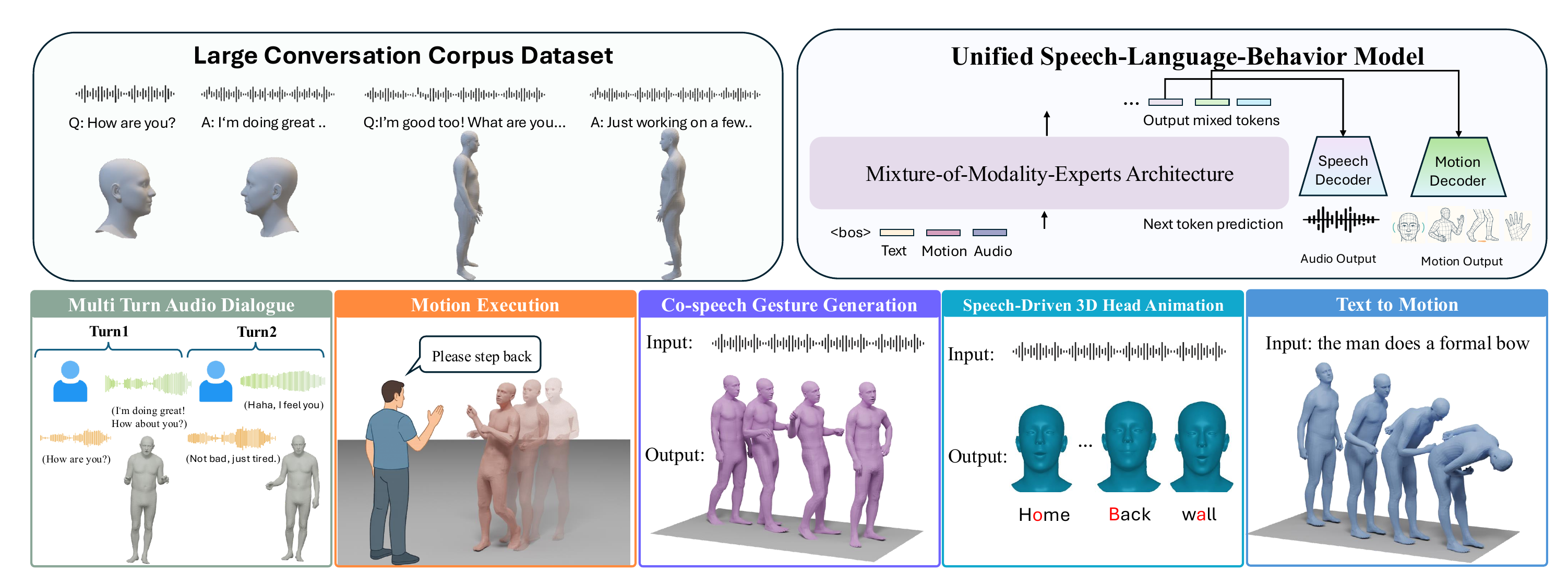}
    \vspace{-0.3in}
    \captionof{figure}{
    We present a novel speech--language--behavior (SLB) model with a mixture--of--modality--experts (MoME) architecture that ingests audio, motion, or text and shares cross-modal information via speech--language--behavior Attention (SLB-Attn). Along with a 1,000-hour conversational behavior corpus that we curate, the model advances toward endowing AI agents with a unified 3D virtual body.}
    \label{fig:teaser}
\end{strip}

\input{sec/0_abstract}
\input{sec/1_intro}
\input{sec/2_relatedwork}
\input{sec/3_method}

\input{sec/4_dataset}
\input{sec/5_experiment}

\input{sec/6_conclusion}

{
    \small
    \bibliographystyle{ieeenat_fullname}
    \bibliography{main}
}
\clearpage
\input{sec/7_supp}

\end{document}

%% file: preamble.tex
\newcommand{\CA}[1]{{\color{teal}#1}}

\renewcommand{\CA}[1]{{\color{black}#1}} 



%% file: sec/0_abstract.tex
\vspace{-0.4in}
\begin{abstract}
\vspace{-0.3in}

Human communication is inherently multimodal and social: words, prosody, and body language jointly carry intent. Yet most prior systems model human behavior as a translation task—co-speech gesture or text--to--motion that maps a fixed utterance to motion clips—without requiring agentic decision-making about when to move, what to do, or how to adapt across multi-turn dialogue. This leads to brittle timing, weak social grounding, and fragmented stacks where speech, text, and motion are trained or inferred in isolation. 
We introduce \textbf{ViBES} (Voice in Behavioral Expression and Synchrony), a conversational 3D agent that jointly plans language and movement and executes dialogue-conditioned body actions. Concretely, ViBES is a speech--language--behavior (SLB) model with a mixture--of--modality--experts (MoME) backbone: modality-partitioned transformer experts for speech, facial expression, and body motion.
The model processes interleaved multimodal token streams with hard routing by modality (parameters are split per expert), while sharing information through cross-expert attention. By leveraging strong pretrained speech-language models, the agent supports mixed-initiative interaction: users can speak, type, or issue body-action directives mid-conversation, and the system exposes controllable behavior hooks for streaming responses.
We further benchmark on multi-turn conversation with automatic metrics of dialogue–motion alignment and behavior quality, and observe consistent gains over strong co-speech and text-to-motion baselines. ViBES goes beyond “speech-conditioned motion generation” toward agentic virtual bodies where language, prosody, and movement are jointly generated, enabling controllable, socially competent 3D interaction. Code and data will be made available at: {\small\hypersetup{urlcolor=rosepink}\url{ai.stanford.edu/~juze/ViBES/}}

\end{abstract}

%% file: sec/1_intro.tex
\vspace{-0.2in}
\section{Introduction}
\vspace{-0.05in}

Nowadays, chatbots can converse fluently, imitate voices, and perform surprising zero- and few-shot generalization across a wide range of tasks~\cite {radford2018improving, radford2019language, brown2020language, achiam2023gpt, liu2024deepseek, hurst2024gpt, team2023gemini, comanici2025gemini, li2026toward}. These capabilities have pushed the community toward a vision of agentic AI—systems that do more than respond: they plan, take initiative, and orchestrate tools over time~\cite{park2023generative}. While exicting, these powerful “digital brains” typically remain confined to text~\cite{radford2018improving, radford2019language, brown2020language, achiam2023gpt, liu2024deepseek, shen2025fine} and audio~\cite{hurst2024gpt}.
On the other hand, human communication \CA{is not limited to} voice and text -- we also use our facial expression and body gesture to communicate \CA{with others}.
Empowering the agent with a body that can act like humans is thus of vital importance. This also has important applications in robotics~\cite{driess2023palm, brohan2022rt, zitkovich2023rt, black2410pi0, xu2025stare, liu2026palm} where robots need to communicate and react to humans in a human-like way.



A lot of recent work has focused on building conversational agents with either text or audio modality~\cite{zeng2024glm, Qwen2.5-Omni, huang2025step, du2024cosyvoice, du2024cosyvoice2, du2025cosyvoice3, defossez2024moshi, higgsaudio2025, zhang2023neuraldome, zhao2024m, zhang2024hoi, zhang2023ikol, zhang2022mutual, fu2025learning} by training on thousands of hours of audio-text data to produce high-quality conversation abilities.
Concurrently, there are two other lines of work that attempt to build conversational agents with a focus on motion:
co-speech gesture generation~\cite{liu2022disco, liu22beat, ijcai2023p650, habibie2021learning, yi23generating, chen2024Synerg, liu24emage, chen2025language, nagy2025gesture} and text-to-motion~\cite{guo2022tm2t, guo2022generating, chen2023executing, zhang2023generating, jiang2023motiongpt, lou2023diversemotion, lou2023diversemotion, guo2024momask, zhang2025motion, tevet2023human, tevet2024closd, lin2025quest}. Both largely frame behavior as modality translation (e.g., audio/text to motion sequences via diffusion~\cite{chen2023executing, tevet2023human} or Transformer models~\cite{jiang2023motiongpt, wang2024motiongpt} on SMPL~\cite{SMPL:2015, SMPL-X:2019, MANO:SIGGRAPHASIA:2017} or human skeletons~\cite{guo2022generating}), optimizing rhythmic entrainment and kinematics.
While these methods can render human motion following some conditioning signals, they do not require behavior intelligence or agentic decision-making when the input prompt requires reasoning. How can we build a model that interfaces multi-modalities, reasons, and also generates human-like behaviors?


A seemingly promising workaround is to bolt a speech LLM onto a motion generator~\cite{chen2025language,jiang2025solami}. In practice, this \CA{intuitive two-stage approach} struggles: there is no unified policy over timing and selection, no shared conversational state, and thus no guarantee of cross-turn consistency or safety. The \CA{most relevant recent work}, LoM~\cite{chen2025language} and SOLAMI~\cite{jiang2025solami}, primarily align speech with body motion under limited data; they emphasize modality alignment rather than preserving conversational intelligence. A true 3D conversational agent should not only produce co-speech gestures while answering, but also follow explicit action instructions (e.g., “Could you step back and wave?”)—capabilities existing systems \CA{do not currently have}.

To address this issue, we present ViBES, a speech--language--behavior (SLB) model with a mixture--of--modality--experts (MoME) architecture that ingests audio, motion, or text and shares cross-modal information via speech--language--behavior attention (SLB-Attn) (Fig.~\ref{fig:teaser}). We \CA{utilize} a strong pretrained audio–text backbone and interleave the newly added modality through an audio–text–motion pathway within a global cross-attention layer that operates over an interleaved token stream. Crucially, our approach does not require large-scale audio–text–motion pretraining. Instead, we leverage the pretrained capacity of speech LLMs~\cite{zeng2024glm} and attach a lightweight per-layer modality expert—a small Transformer block that produces face and body queries and reads the backbone’s key/value via cross-attention. Because these experts are side-car modules, the backbone’s architecture and weights remain intact, allowing us to \CA{utilize} off-the-shelf checkpoints while adding new modalities with minimal overhead. The design is “sparse” in that cross-modal interaction is restricted to attention reads (no heavy feature concatenation or write-back), and it scales by duplicating the expert across layers without retraining the language/speech core. 

Training this model requires large-scale, time-aligned audio–text–motion datasets.
However, today’s datasets are limited~\cite{liu24emage,mclean2025embody,guo2022generating,AMASS:ICCV:2019}: we typically have only pairwise alignments (text-to-motion, audio-to-motion) rather than large-scale, time-synchronized audio–text–motion supervision. 
To overcome the data bottleneck,
we curate a 1{,}000-hour dataset from complementary sources: in-the-wild YouTube conversation videos and existing motion datasets~\cite{AMASS:ICCV:2019,guo2022generating,liu24emage,sun2024diffposetalk, reece2023candor}. The YouTube portion spans interviews, podcasts, talks, and casual dialogues. For each video, we automatically recover monocular 3D human motion—body and hands when visible—and facial expressions using a single-view pipeline, then temporally align these signals with the speech and transcripts. The resulting corpus provides dense, time-synchronized audio–text–motion triplets, enabling, for the first time at this scale, large-scale training of a conversational agent’s body alongside its language and speech.

To evaluate our trained model, we introduce a new benchmark that jointly assesses dialogue understanding, social appropriateness, and motion quality, leveraging multimodal LLM judges~\cite{hurst2024gpt} together with task-specific metrics to cover the full spectrum of conversational behavior. We compare ViBES against a strong state-of-the-art system and several baselines and show that it outperforms prior methods across all metrics. We further evaluate on co-speech gesture generation and text-to-speech benchmarks~\cite{liu24emage, guo2022generating}, where ViBES also achieves state-of-the-art results.

In summary, we present a unified sparse multi-modal model that integrates an unfrozen motion stack with a pretrained audio–text backbone via latent cross-attention, enabling dialogue-conditioned motion while preserving linguistic and prosodic intelligence. We also introduce a large-scale dataset of over 1{,}000 hours of synchronized audio–text–motion interactions and a benchmark suite evaluating dialogue–motion alignment, timing, kinematics, and social appropriateness. The model outperforms strong baselines in responsiveness, motion alignment, and social competence. We believe reframing non-verbal behavior as a core component of conversational intelligence and advancing toward socially adept 3D agents that communicate through language, voice, and movement.

%% file: sec/2_relatedwork.tex
\vspace{-0.1in}
\section{Related Work}
\vspace{-0.05in}
\subsection{Conversational Avatar Generation}
\vspace{-0.05in}
A conversational avatar fundamentally relies on two key components: a \textit{speech module} to generate natural voice, and a \textit{visual module} to synthesize realistic motion, including facial expressions and gestures, in either 2D or 3D form.

Speech dialog research splits into two paradigms: modular and end-to-end. Modular systems pair an LLM with a TTS/vocoder—LLM emits text or semantic tokens, TTS renders waveforms~\cite{du2024cosyvoice,du2024cosyvoice2,du2025cosyvoice3,lyu2025build,defossez2024moshi, yu2024salmonn, peng2025vibevoice}. End-to-end voice agents integrate understanding and speech generation in a single model, tightly coupling semantics, prosody, and emotion~\cite{zeng2024glm,huang2025step,chu2024qwen2}.

In visual generation~\cite{jiang2025omnihuman, lin2025omnihuman, li2025infinityhuman, chen2025talkcuts, ma2025follow, yang2025videogen}, particularly 2D portrait animation, approaches are commonly grouped into video-driven and audio-driven formulations. Video-driven methods transfer motion from reference clips using explicit or implicit keypoints~\cite{guo2024liveportrait,xu2025hunyuanportrait,wang2021facevid2vid}. Audio-driven methods synthesize talking heads directly from speech, either end-to-end (e.g., Loopy~\cite{jiang2024loopy}, Emo~\cite{tian2024emo}) or via two-stage designs (e.g., VASA-1~\cite{xu2024vasa}), with related variants~\cite{wei2024aniportrait,xie2025x}. 
Despite scaling gains, 2D representations limit applicability to interactive robotics and game engines and often degrade under viewpoint changes. 
Beyond 2D, 3D avatar generation boosts expressiveness: audio-to-face~\cite{nocentini2024scantalk,lee2019talking, chu2025artalk,fan2024unitalker,peng2023selftalk,sung2024multitalk, sun2024diffposetalk, danvevcek2025supervising} models target synchronization, and gesture generation emphasizes rhythmic and semantic alignment. Most works tackle face or gestures in isolation, whereas we aim to unify speech, facial expressions, and full-body motion.

\vspace{-0.05in}
\subsection{Multi-modal Motion Generation}
\vspace{-0.05in}

Human motion generation can be conditioned on text~\cite{zhang2025motion,jiang2023motiongpt,chen2023executing, wu2024mote,hou2025motionverse,zhu2025motiongpt3, guo2024momask, guo2022generating, zhang2024both2hands, yu2025socialgen}, audio~\cite{liu2022disco, liu22beat, ijcai2023p650, habibie2021learning, yi23generating, chen2024Synerg, liu24emage,chen2025language, zhang2025social, liu2025mimicparts}, music~\cite{li2021aistpp, siyao2022bailando, liu2025mosa, tseng2023edge, li2023finedance, huang2024beat, liu2025gcdance, wang2025dancechat, ghosh2025duetgen, liu2025dgfm}, and increasingly on multi-modal control~\cite{zhang2024lmm,li2025genmo,unimumo2024, xu2025omnimotion, liang2024llava, cong2023weakly}. Text-driven methods are well studied, including diffusion-based models~\cite{tevet2023human,chen2023executing}, transformer-based autoregressive models~\cite{jiang2023motiongpt,zhu2025motiongpt3, zhang2024motiongpt, wang2024motiongpt} and hybrid diffusion–transformer designs~\cite{zhu2025motiongpt3}. Audio-driven methods often address audio-to-motion synthesis, generating full-body motion from speech with an emphasis on rhythmic alignment~\cite{liu24emage,chen2025language}. 
%
Recent work advances unified conditioning, enabling single models to respond to multiple signals (e.g., GenMo~\cite{li2025genmo}, LMM~\cite{zhang2024lmm}, RapVerse~\cite{chen2025rapverse}, LoM~\cite{chen2025language}). The closest, SOLAMI~\cite{jiang2025solami}, aligns speech with body motion but lacks explicit text conditioning and an LLM reasoning backbone, and omits expressive facial motion. Building on this line, our framework seeks to unify multimodal control for avatar generation, combining audio, text, and conversational reasoning into a single model with question-answering capability.

\vspace{-0.05in}
\subsection{Multimodal Large Language Models}
\vspace{-0.05in}

Recent MLLMs highlight unified modeling across text, images, and audio~\cite{liang2024mixture,  team2024chameleon, li2023blip, zhao2024image, du2026unsupervised}. Two fusion patterns dominate: single-stream models that concatenate multimodal tokens into one Transformer, and dual-stream/two-tower models that keep modality-specific backbones with cross/co-attention (e.g., ViLBERT~\cite{lu2019vilbert}, LXMERT~\cite{tan2019lxmert}). In generative vision, DiT and multimodal variants are now standard; Stable Diffusion 3 adopts MMDiT with separate image/language weights and joint attention for efficient bidirectional exchange\cite{peebles2023scalable,esser2024scaling}. A complementary approach inserts gated cross-attention into a (partially) frozen LM to condition on visual tokens without full retraining~\cite{alayrac2022flamingo}.

Beyond dense fusion, expert-based designs scale by splitting parameters by modality and routing compute. MoME layers in VLMo and LIMoE select modality-aware experts via a router, yielding strong efficiency/accuracy trade-offs~\cite{bao2022vlmo,mustafa2022multimodal}. Mixture-of-Transformers~\cite{liang2024mixture} decouples non-embedding parameters per modality while retaining global self-attention over mixed sequences, matching dense baselines with far fewer FLOPs. In robotics, VLA models link perception and policy via multimodal attention/adapters (e.g., RT-2, OpenVLA)~\cite{zitkovich2023rt,kim2024openvla,won2025dual}. Inspired by these trends, we adopt parameter-split Transformers with cross-modal fusion inside attention, supporting both autoregressive and diffusion generation for efficient pretraining and high-fidelity, controllable avatar synthesis.

%% file: sec/3_method.tex
\vspace{-0.2in}
\section{Architecture}
\label{sec:method}
\vspace{-0.1in}

\begin{figure*}[t]
    \centering
    \includegraphics[width=0.9\linewidth]{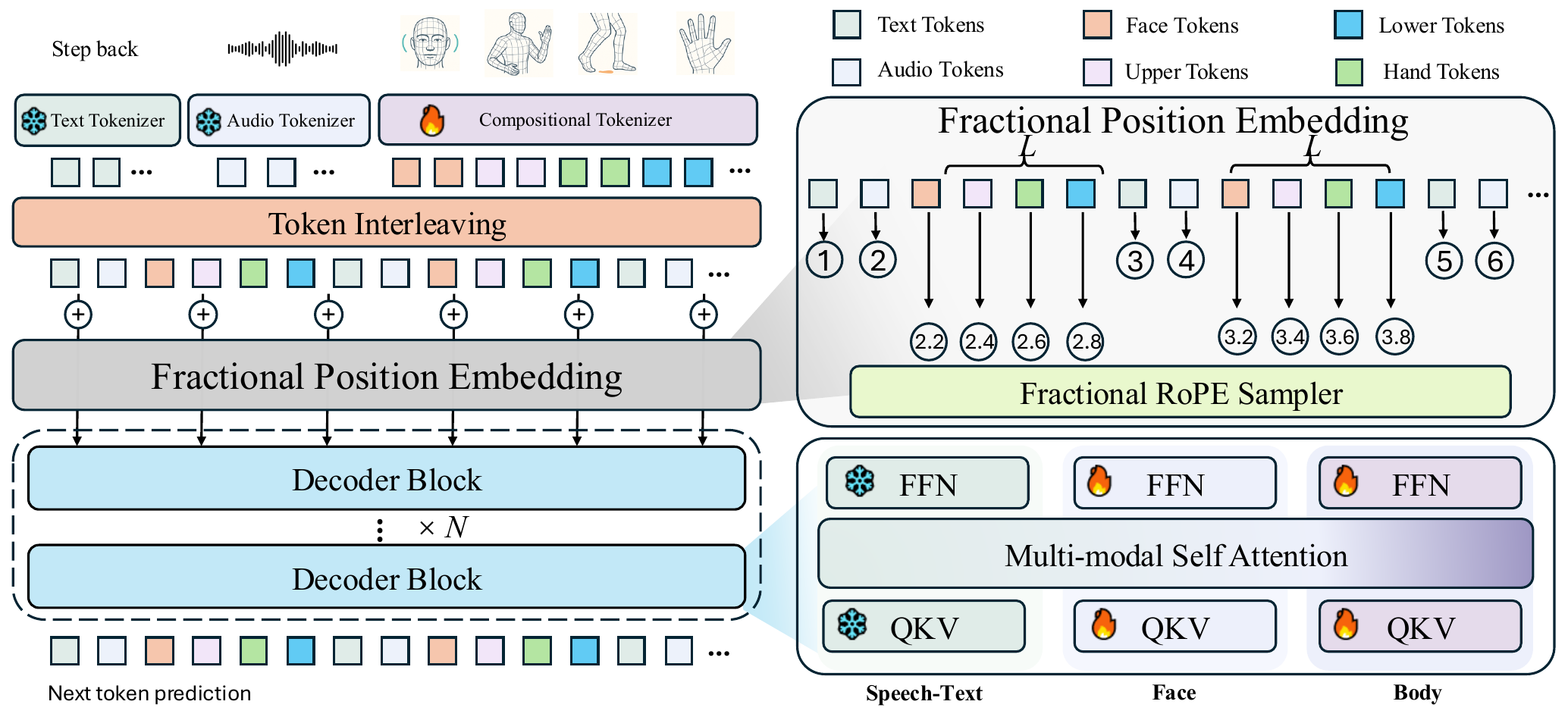}
    \vspace{-0.1in}
    \caption{Model overview. The model adopts an autoregressive structure that converts all modalities into a unified token space. It consists of a Speech–Language–Behavior model with a Mixture-of-Modality-Experts (MoME) architecture, which processes audio, motion, and text inputs while sharing cross-modal information through the proposed Speech–Language–Behavior Attention (SLB-Attn) mechanism.}
    \label{fig:approach}
    \vspace{-0.1in}
\end{figure*}

In this section, we present ViBES, an end-to-end architecture for a 3D conversational agent (Fig.~\ref{fig:approach}). ViBES is an SLB model with MoME: three Transformer experts—a speech--language expert for understanding and generation, a facial-expression expert, and a body-motion expert—coupled via cross-modal attention to enforce speech, face and body synergy.
Accordingly, ViBES comprises three main components: (i) a multimodal tokenizer that converts each sequential modality into multi-stream tokens while faithfully preserving fine-grained audio and motion dynamics; (ii) an LLM backbone that jointly models these token streams under the next-token prediction paradigm and exchanges information via SLB-Attn; and (iii) modality-aware fractional rotary positional embeddings (RoPE) that precisely encode cross-modal time alignment and relative positions.

\vspace{-0.05in}
\subsection{Modality Representations \& Tokenizaion}
\label{sec:notation}

Precise cross-modal time alignment is critical for speaking agents. In practice, speech LLMs discretize audio at 12.5~\cite{zeng2024glm}, 25~\cite{du2024cosyvoice} or 50\,fps~\cite{hsu2021hubert}, whereas motion streams are typically 20~\cite{zhang2023generating} or 30\,fps~\cite{liu24emage}, inducing systematic misalignment. We therefore standardize on a 25\,fps universal clock: audio tokens are produced at 12.5\,fps and all motion streams are resampled to 25\,fps before fusion.
We utilize three types of modality tokenization to convert each modality into a discrete token space. 

\tightpara{Text and speech tokenization.}
We use Tiktoken, from GLM-4-9B~\cite{glm2024chatglm} model, as text tokenizer to produce tokens $X=\{x_t\}_{t=1}^{T}$ (sub-word tokens with word timestamps for temporal alignment) where $T$ is the window length. For speech tokenization, the goal is two-fold: to preserve high-precision acoustic tokens with enough information to reconstruct the original audio and to preserve semantic information that captures the speaker's emotion. Here, we adopt the 12.5Hz speech tokenizer variant from GLM-4-voice~\cite{zeng2024glm}. The speech tokenizer was fine-tuned on a pretrained automatic speech whisper-large-v3 in the Whisper family~\cite{radford2023robust} with an additional pooling layer and a vector quantization layer in the middle of the encoder.

\vspace{-0.05in}
\tightpara{Face and body tokenization.}
We represent facial motion with the 3D morphable model FLAME~\cite{FLAME:SiggraphAsia2017} and body motion with the parametric model SMPL-X~\cite{SMPL-X:2019}. For each frame of the speaking sequence, the face is parameterized by shape $\beta_f\!\in\!\mathbb{R}^{100}$, expression $\psi_f\!\in\!\mathbb{R}^{100}$, and a 6D pose vector $\theta_f\!\in\!\mathbb{R}^{6}$ (continuous rotation); the body is parameterized by shape $\beta_b\!\in\!\mathbb{R}^{300}$ and joint poses $\theta_b\!\in\!\mathbb{R}^{55}$. 
Two representation conventions are prevalent in the community: the \emph{skeletal} representation popularized by HumanML3D~\cite{guo2022generating} and the \emph{rotation-based} representation for expressive motion~\cite{liu24emage}. These choices hinder direct cross-paper comparison. To ensure fairness, we export motion in \emph{both} formats and tokenize them accordingly—(i) skeletal sequences following HumanML3D and (ii) rotation-space sequences, each aligned to our 25\,fps clock and used for evaluation against their respective baselines. For faces, we adopt the facial tokenizer introduced in LoM~\cite{chen2025language} and fine-tune it on our data without latent downsampling, yielding 25\,fps face tokens; for expressive body motion, we use compositional tokenizer (upper body, lower body, hands) without additional fine-tuning and apply $\times4$ temporal downsampling, resulting in 6.25\,fps body tokens; for the HumanML3D setting, we follow MotionGPT~\cite{jiang2023motiongpt}, which likewise produces 6.25\,fps tokens. All streams are aligned to a 25\,fps master clock via a token–frame map for multimodal fusion.

\vspace{-0.05in}
\subsection{Speech--Language--Behavior Model}
\label{sec:arch}

ViBES adopts a Mixture-of-Modality-Experts (MoME) with three modality-specific FFN experts and LayerNorms: a text--speech(TS) expert operating on text and speech tokens for language comprehension and speech generation, a facial expert for facial-expression synthesis, and a body expert for upper, lower body, and hands. Unlike sparse MoE architectures with learned routers~\cite{shazeer2017outrageously}, we use hard routing that deterministically assigns tokens to experts by modality. Tokens are interleaved in the sequence. Each expert has its own FFN and LayerNorm and we employ modality-specific projections $(W^{Q,K,V,O}_{\mathrm{ts}}, W^{Q,K,V,O}_{\mathrm{face}}, W^{Q,K,V,O}_{\mathrm{body}})$; attention is computed only over allowed modality pairs. Our architecture is heavily motivated by similar architectures from  MOT~\cite{liang2024mixture}, MMDit~\cite{esser2024scaling}, and BAGEL~\cite{deng2025emerging}. Let the \emph{interleaved} sequence be $\mathbf{x}=(x^{m_1}_{1},\ldots,x^{m_T}_{T})$ with modality labels $m_t\in\mathcal{M}=\{\text{text},\text{speech},\text{face},\text{upper},\text{lower},\text{hands}\}$; layer input $\mathbf{h}^{(\ell)}\in\mathbb{R}^{T\times d}$ (batch omitted), and $\mathbf{h}^{(0)}$ the embedding of $\mathbf{x}$. Define subsets
$
\mathcal{M}_{\mathrm{ts}}=\{\text{text},\text{speech}\},
\mathcal{M}_{\mathrm{face}}=\{\text{face}\},
\mathcal{M}_{\mathrm{body}}=\{\text{upper},\text{lower},\text{hands}\}.
$

\tightpara{Speech and text branch.}
Only text and speech tokens route to this expert, which has its own $\mathrm{LN}_{\mathrm{ts}}$ and $\mathrm{FFN}_{\mathrm{ts}}$ (initialized from GLM-4-Voice and frozen). TS queries attend only to TS keys and values with TS-specific projections:
\begin{equation}
\label{eq:ts-attn-proj}
\begin{aligned}
\mathbf{Q}_{\mathrm{ts}}=\mathrm{LN}_{\mathrm{attn}}(\mathbf{h}^{(\ell)}_{\mathrm{ts}})W^{Q}_{\mathrm{ts}},\\
\mathbf{K}_{\mathrm{ts}}=\mathrm{LN}_{\mathrm{attn}}(\mathbf{h}^{(\ell)}_{\mathrm{ts}})W^{K}_{\mathrm{ts}},\\
\mathbf{V}_{\mathrm{ts}}=\mathrm{LN}_{\mathrm{attn}}(\mathbf{h}^{(\ell)}_{\mathrm{ts}})W^{V}_{\mathrm{ts}}.
\end{aligned}
\end{equation}
The TS attention update is
\begin{equation}
\label{eq:ts-attn}
\tilde{\mathbf{h}}^{(\ell)}_{\mathrm{ts}}
=\mathbf{h}^{(\ell)}_{\mathrm{ts}}
+\mathrm{Softmax}\!\Big(\tfrac{\mathbf{Q}_{\mathrm{ts}}\mathbf{K}_{\mathrm{ts}}^{\top}}{\sqrt{d_h}}\Big)\,\mathbf{V}_{\mathrm{ts}}\,W^{O}_{\mathrm{ts}}.
\end{equation}
Then the TS expert applies its \emph{own} normalization and FFN:
\begin{equation}
\label{eq:ts-ffn}
\mathbf{h}^{(\ell+1)}_{t}
=\tilde{\mathbf{h}}^{(\ell)}_{t}
+\mathrm{FFN}_{\mathrm{ts}}\!\big(\mathrm{LN}_{\mathrm{ts}}(\tilde{\mathbf{h}}^{(\ell)}_{t})\big), \text{for } m_t\in\mathcal{M}_{\mathrm{ts}}.
\end{equation}

\tightpara{Face and body branch.}
Facial tokens route to $(\mathrm{LN}_{\mathrm{face}},\mathrm{FFN}_{\mathrm{face}})$; body tokens (upper, lower and hands) route to $(\mathrm{LN}_{\mathrm{body}},\mathrm{FFN}_{\mathrm{body}})$. Queries use \textbf{their own} projections, but keys/values come \emph{only} from the TS stream (no Face$\leftrightarrow$Body attention):
\begin{equation}
\label{eq:face-body-proj}
\begin{aligned}
\mathbf{Q}_{\mathrm{face}}=\mathrm{LN}_{\mathrm{attn}}(\mathbf{h}^{(\ell)}_{\mathrm{face}})W^{Q}_{\mathrm{face}},\\
\mathbf{Q}_{\mathrm{body}}=\mathrm{LN}_{\mathrm{attn}}(\mathbf{h}^{(\ell)}_{\mathrm{body}})W^{Q}_{\mathrm{body}}.
\end{aligned}
\end{equation}
Facial and body attention updates (to TS keys/values) are
\begin{equation}
\label{eq:face-body-attn}
\begin{aligned}
\tilde{\mathbf{h}}^{(\ell)}_{\mathrm{face}}
=\mathbf{h}^{(\ell)}_{\mathrm{face}}
+\mathrm{Softmax}\!\Big(\tfrac{\mathbf{Q}_{\mathrm{face}}\mathbf{K}_{\mathrm{ts}}^{\top}}{\sqrt{d_h}}\Big)\,\mathbf{V}_{\mathrm{ts}}\,W^{O}_{\mathrm{face}}, \\
\tilde{\mathbf{h}}^{(\ell)}_{\mathrm{body}}
=\mathbf{h}^{(\ell)}_{\mathrm{body}}
+\mathrm{Softmax}\!\Big(\tfrac{\mathbf{Q}_{\mathrm{body}}\mathbf{K}_{\mathrm{ts}}^{\top}}{\sqrt{d_h}}\Big)\,\mathbf{V}_{\mathrm{ts}}\,W^{O}_{\mathrm{body}}.
\end{aligned}
\end{equation}
Finally, each expert applies its own normalization and FFN to its positions (tokens remain interleaved in the global sequence):
\begin{equation}
\label{eq:face-body-ffn}
\mathbf{h}^{(\ell+1)}_{t}
=
\begin{cases}
\tilde{\mathbf{h}}^{(\ell)}_{t}
+\mathrm{FFN}_{\mathrm{face}}\!\big(\mathrm{LN}_{\mathrm{face}}(\tilde{\mathbf{h}}^{(\ell)}_{t})\big),
& \text{if } m_t\in\mathcal{M}_{\mathrm{face}},\\[4pt]
\tilde{\mathbf{h}}^{(\ell)}_{t}
+\mathrm{FFN}_{\mathrm{body}}\!\big(\mathrm{LN}_{\mathrm{body}}(\tilde{\mathbf{h}}^{(\ell)}_{t})\big),
& \text{if } m_t\in\mathcal{M}_{\mathrm{body}}.
\end{cases}
\end{equation}

It is worth noting that Eqs.~\eqref{eq:ts-attn-proj}--\eqref{eq:face-body-attn} employ modality-specific projections: TS attends to TS, whereas facial/body queries attend only to TS. In ablations, enabling face and body attention yields no measurable improvement; once conditioned on TS, the face and body streams appear largely independent. We therefore disable face and body attention by design; see Ablation Study~\ref {sec:experiment} for more details.

\subsection{Multi-modal Fractional RoPE}
\label{sec:rope}
Interleaved-token mixed training has been shown to efficiently align modalities—especially for sequential data~\cite{alayrac2022flamingo, zeng2024glm, team2024chameleon, zhou2024transfusion}. In conversational agents, each modality is inherently sequential and thus requires precise cross-modal synchronization. Thus, we encode interleaved multi-modal tokens on a \emph{single} rotary timeline by assigning each token a scalar (possibly non-integer) index $s_t\in\mathbb{R}$ that preserves cross-modal timing. The timeline is anchored by the text--speech (TS) stream, consistent with our attention topology where keys/values originate from TS.

\tightpara{Setup and anchors.}
Let the interleaved sequence be $\{x^{m_t}_t\}_{t=1}^{T}$ with $m_t\in\mathcal{M}=\{\text{text},\text{speech},\text{face},\text{upper},\text{lower},\text{hands}\}$. Denote the TS subset $\mathcal{M}_{\mathrm{ts}}=\{\text{text},\text{speech}\}$ and the motion subset $\mathcal{M}_{\mathrm{mo}}=\mathcal{M}\setminus\mathcal{M}_{\mathrm{ts}}$. Let $\mathcal{A}=\{a_0<\dots<a_{L-1}\}$ be the indices of tokens whose modality lies in $\mathcal{M}_{\mathrm{ts}}$ (TS anchors). Each anchor receives an \emph{integer} index
\[
s_{a_i}=i,\qquad i=0,\dots,L-1.
\]
Assume each token carries a wall-clock timestamp $u(t)$ (e.g., audio frame time; text aligned by ASR/forced alignment). Let $\Delta_i=u(a_{i+1})-u(a_i)$ be the anchor interval length and $\bar{\Delta}$ the median of $\{\Delta_i\}$ for extrapolation.

\tightpara{Fractional indices for motion tokens.}
For a motion token at position $t$ with $m_t\in\mathcal{M}_{\mathrm{mo}}$, find $i$ such that $u(a_i)\le u(t)<u(a_{i+1})$. Its fractional index is the linear interpolation between the surrounding TS anchors:
\begin{equation}
\label{eq:frac-interp}
\alpha_t=\frac{u(t)-u(a_i)}{u(a_{i+1})-u(a_i)}\in[0,1),\qquad
s_t = s_{a_i} + \alpha_t.
\end{equation}
If $u(t)<u(a_0)$ (left tail), use a left extrapolation with step $\bar{\Delta}$:
\begin{equation}
\label{eq:left-tail}
s_t = s_{a_0} - \frac{u(a_0)-u(t)}{\bar{\Delta}}.
\end{equation}
If $u(t)\ge u(a_{L-1})$ (right tail), use a right extrapolation:
\begin{equation}
\label{eq:right-tail}
s_t = s_{a_{L-1}} + \frac{u(t)-u(a_{L-1})}{\bar{\Delta}}.
\end{equation}
This construction yields a globally ordered sequence of indices $\{s_t\}_{t=1}^T$ in which TS positions are integers and motion positions fall at fractional locations aligned to TS time.

\tightpara{Optional per-modality offset and scale.}
To stabilize training with interleaved streams of different frame rates, we optionally apply a modality-specific phase offset $\delta_m\in\mathbb{R}$ and a phase scale $\gamma_m>0$:
\begin{equation}
\label{eq:phase-affine}
\hat{s}_t = \gamma_{m_t}\,s_t + \delta_{m_t}.
\end{equation}
Typical choices set $\gamma_{\mathrm{ts}}=1$, and small offsets for motion (e.g., $\delta_{\mathrm{face}}<0<\delta_{\mathrm{body}}$) to reduce collision at anchor boundaries. When not used, take $\gamma_m=1$ and $\delta_m=0$.

\tightpara{Rotary positional encoding.}
Let $d_{\mathrm{rot}}$ be the rotary dimension (even), and $L=d_{\mathrm{rot}}/2$ the number of frequency channels. With base $b>1$, define the inverse-frequency ladder
\begin{equation}
\label{eq:ladder}
\omega_j = b^{-\,\frac{2j}{d_{\mathrm{rot}}}},\qquad j=0,\dots,L-1.
\end{equation}
Given the (possibly fractional) index $\hat{s}_t$ from \eqref{eq:phase-affine}, the phase for channel $j$ is $\theta_{t,j}=\hat{s}_t\,\omega_j. $
Denote by $\operatorname{rot}(\cdot\,;\theta)$ the 2D rotation applied to coordinate pairs $(2j,2j{+}1)$ of queries/keys. For a query $\mathbf{q}_t$ and a key $\mathbf{k}_i$, RoPE applies
\begin{equation}
\label{eq:rope-apply}
\mathbf{q}_t^{\mathrm{rope}}=\operatorname{rot}(\mathbf{q}_t;\theta_{t,:}),\qquad
\mathbf{k}_i^{\mathrm{rope}}=\operatorname{rot}(\mathbf{k}_i;\theta_{i,:}),
\end{equation}
after which attention uses the standard dot-product with these rotated vectors. Stacking phases in matrix form with $\boldsymbol{\omega}=[\omega_0,\dots,\omega_{L-1}]^\top$ and $\hat{\mathbf{s}}=[\hat{s}_1,\dots,\hat{s}_T]^\top$ gives $\Theta=\hat{\mathbf{s}}\,\boldsymbol{\omega}^{\top}\in\mathbb{R}^{T\times L}$.

%% file: sec/4_dataset.tex
\section{The Converse3D Dataset}
\label{sec:dataset}

\begin{table}[t]
\centering
\caption{Comparison of Converse3D with existing datasets. We report the duration of Converse3D as N/A, since it is constructed by combining web YouTube videos and other datasets, making a direct comparison in hours less meaningful.}
\vspace{-0.1in}
\resizebox{0.48\textwidth}{!}{
\begin{tabular}{lccccccc}
\hline
Datasets & \makecell{Duration \\(hours)} & Q\&A & Facial  & \makecell{Upper\\Body} & \makecell{Lower\\Body} & Mesh & Description \\
\hline
TED~\cite{yoon2020speech}  & 106.1     & \ding{55}& \ding{55} & \ding{51} & \ding{55} & \ding{55} & \ding{55}   \\
TED-Ex~\cite{liu2022learning}   & 100.8 & \ding{55} & \ding{55} & \ding{51} & \ding{55} & \ding{55} & \ding{55}   \\
EGGS~\cite{ghorbani2023zeroeggs} & 2 & \ding{55}  & \ding{55} & \ding{51} & \ding{51} & \ding{55} & \ding{55}  \\
BEAT~\cite{liu2022beat}     & 76  & \ding{55}    & \ding{55} & \ding{51} & \ding{51} & \ding{51} & \ding{55}  \\
SHOW~\cite{yi2023generating}  & 26.9  & \ding{55} & \ding{51} & \ding{51} & \ding{55} & \ding{51} & \ding{55}   \\
TWH16.2~\cite{lee2019talking}   & 17  & \ding{55}   & \ding{55}  & \ding{51}  & \ding{51}  & \ding{51} & \ding{55} \\
BEAT2~\cite{liu24emage}    & 60   & \ding{55}  & \ding{51} & \ding{51} & \ding{51} & \ding{51} & \ding{55}  \\
DND~\cite{mughal2024convofusion}  & 6   & \ding{55}   & \ding{55} & \ding{51} & \ding{51} & \ding{55} & \ding{55}  \\
Photoreal~\cite{ng2024audio}  & 8   & \ding{55}  & \ding{51} & \ding{51} & \ding{51} & \ding{55} & \ding{55} \\
GES-Inter~\cite{qi2025co}  & 8  & \ding{51}    & \ding{51} & \ding{51} & \ding{55} & \ding{51} & \ding{55}  \\
Embody3D~\cite{mclean2025embody} & 500 & \ding{51}    & \ding{55} &  \ding{51} & \ding{51}  & \ding{51}  & \ding{51} \\
Seamless Interaction~\cite{agrawal2025seamless} & \textbf{4,065} & \ding{51}    & \ding{55} &  \ding{51} & \ding{55}  & \ding{51}  & \ding{55} \\
Converse3D (Ours) & NA   & \ding{51} & \ding{51} & \ding{51} & \ding{51} & \ding{51} & \ding{51} \\
\hline
\end{tabular}
}
\vspace{-0.1in}
\end{table}

A significant bottleneck for developing 3D conversational agents is the lack of large-scale, time-aligned audio–text–motion datasets. Existing datasets are typically fragmented, providing only pairwise alignments(e.g., text-to-motion, text-to-speech) and limited coverage. To overcome this, we introduce Converse3D, a corpus of roughly 1{,}000 hours designed for 3D conversational agents, curated from three complementary sources: (i) in-the-wild YouTube conversations spanning interviews, podcasts, talks, and casual dialogues; (ii) high-quality motion datasets with text or audio annotations; and (iii) carefully controlled synthesis to complete missing modalities. All motion is represented in SMPL-X~\cite{SMPL-X:2019} and FLAME \cite{FLAME:SiggraphAsia2017}, and different streams (text, face, body, hands) are temporally aligned to a unified 25\,fps clock to support streaming training and inference.

\subsection{Data Collection from the YouTube Corpus}
We curate long-form conversational videos (interviews, roundtables, talks with audience interaction) and process them with a unified pipeline. Voice activity detection~\cite{tao2021someone}, word-level ASR~\cite{radford2023robust}, and speaker diarization segment multi-party audio into utterance windows. For each window, we estimate SMPL-X body and hands and FLAME face parameter representations using robust monocular methods~\cite{goel2023humans,pavlakos2024reconstructing,filntisis2022visual}, then resample all streams to 25 fps with per-frame timestamps and token–frame maps. Quality control filters low-confidence geometry (high reprojection error, self-intersection), enforces visibility checks before completion, and verifies audio/text integrity (SNR, WER) and temporal alignment consistency. 

\subsection{Consolidating Existing Motion Datasets}


While most motion datasets cover only subsets of modalities, their scale and fidelity are valuable for cross-modal learning. We integrate HumanML3D ($\sim$24K text–motion pairs), BEAT2 ($\sim$70 hours audio–motion), Embody3D ($\sim$50 hours audio with full body), and TFHP (audio with facial annotations). All sequences are retargeted to a common SMPL-X/FLAME parameters, then resampled to 25 fps.

\subsection{Synthesizing Tri-Modal Supervision from Fragmented Sources}

To close residual modality gaps, we synthesize missing signals conservatively. ARTalk~\cite{chu2025artalk} refines lip sync on a small subset, producing FLAME trajectories aligned to audio; LoM~\cite{chen2025language} completes occluded hands only to avoid altering global dynamics. When only motion and a caption are available, GPT-4o~\cite{hurst2024gpt} generates situational prompts that we render with a high-fidelity TTS, yielding paired audio–text queries aligned with the motion. 

%% file: sec/5_experiment.tex
\section{Experiments}
\label{sec:experiment}
In this section, we first introduce a new benchmark to evaluate agentic conversational behavior for ViBES and several baselines. We then demonstrate the versatility of our model on several downstream tasks: talking-head synthesis, co-speech gesture generation, and text-to-motion, and compare against state-of-the-art methods on existing benchmarks.

\begin{table*}[t]
\setlength{\tabcolsep}{5pt}
\centering
\caption{Benchmarking conversational behavior. ``$\uparrow$'' indicates that higher values are better.
The best results are in \textbf{bold}.}
\vspace{-0.1in}
\resizebox{0.8\textwidth}{!}{
\begin{tabular}{lcccccccc}
\toprule
\multirow{1}{*}{Methods} 
 & Input &\makecell{ R1-Balanced $\downarrow$} & \makecell{R1-Motion$\downarrow$} & \makecell{ R1-Conv$\downarrow$ } & \makecell{ R3-Balanced$\downarrow$ }  & \makecell{ MM Dist$\downarrow$ } & \makecell{ FID$\downarrow$ } & \makecell{ Div$\rightarrow$ } \\
\midrule

Ground Truth & NA & 0.712 & 0.546 & 0.999 & 0.883 & 2.342 & 0.0 & 11.05 \\

\midrule

\multicolumn{9}{c}{\textit{Co-speech gesture generation methods}} \\
SynTalker~\cite{chen2024Synerg} & Response audio & 0.288 & 0.025 & 0.744 & 0.390 & 3.650 & 406.3 & 1.70\\
EMAGE~\cite{liu24emage}  & Response audio & 0.258 & 0.025 & 0.663 & 0.378 & 3.639 & 371.7 & 2.37 \\
LOM~\cite{chen2025language}  & Response audio & 0.323 & 0.025 & 0.621 & 0.498 & 3.435 & 373.8 & 2.49 \\
\midrule
\multicolumn{9}{c}{\textit{Text-to-motion generation methods}} \\
T2M~\cite{guo2022generating} & Question text& 0.223 & 0.180 & 0.297 & 0.476 & 3.682 & 306.2 & 6.14 \\
MotionGPT~\cite{jiang2023motiongpt} & Question text& 0.244 & 0.123 & 0.452 & 0.475 & 3.734 & 262.2 & 7.40 \\
MoMask~\cite{guo2024momask}   & Question text & 0.293 & 0.159 & 0.524 & 0.528 & 3.675 & 265.5 & 7.38 \\
\midrule
\multicolumn{9}{c}{\textit{Speech–Language–Behavior Unified Models}} \\
Ours & Question text & \textbf{0.467} & \textbf{0.308} &  \textbf{0.745} &  \textbf{0.671} &  \textbf{3.178} &  \textbf{93.9} &  \textbf{10.73}  \\
\bottomrule
\end{tabular}}
\label{tab:quantitative_behavior}
\end{table*}

\subsection{Benchmark for Conversational Behavior}

Evaluating conversational motion generation is challenging: body movement during speech is inherently one-to-many, and standard motion-text retrieval metrics conflate two distinct capabilities—producing semantically specific actions (e.g., ``wave your right hand'') and generating plausible conversational gestures for open-ended questions (e.g., ``What does happiness mean to you?''). We address this with a balanced R-Precision protocol on the Converse3D test set (4{,}921 samples: 3{,}128 motion-descriptive from AMASS and 1{,}793 conversational from BEAT2). A Question-to-Motion (Q2M) contrastive evaluator, built on a CLIP-style architecture with a Transformer motion encoder and a frozen CLIP~\cite{radford2021learning} text encoder, maps both questions and motions (22 body joints in 6D rotation plus root velocity). Each evaluation batch contains 16 motion-descriptive and 16 conversational samples. For motion-descriptive queries, we apply strict 1-to-1 matching: the model must retrieve the exact paired motion. For conversational queries, we apply relaxed matching: retrieval of any conversational gesture in the batch counts as correct, reflecting the genuine ambiguity of co-speech gesture. We report the balanced R-Precision (weighted average of both types), type-specific R-Precision, Matching Distance, FID against GT motion embeddings, and Diversity.
We report the conversational behavior benchmark in Table~\ref{tab:quantitative_behavior}. ViBES achieves stronger question-to-motion alignment, indicating that joint training within the speech--language--behavior framework provides clear performance gains over prior co-speech gesture generation and text-to-motion baselines.

\begin{figure}[t]
    \centering
    \includegraphics[width=1.0\linewidth]{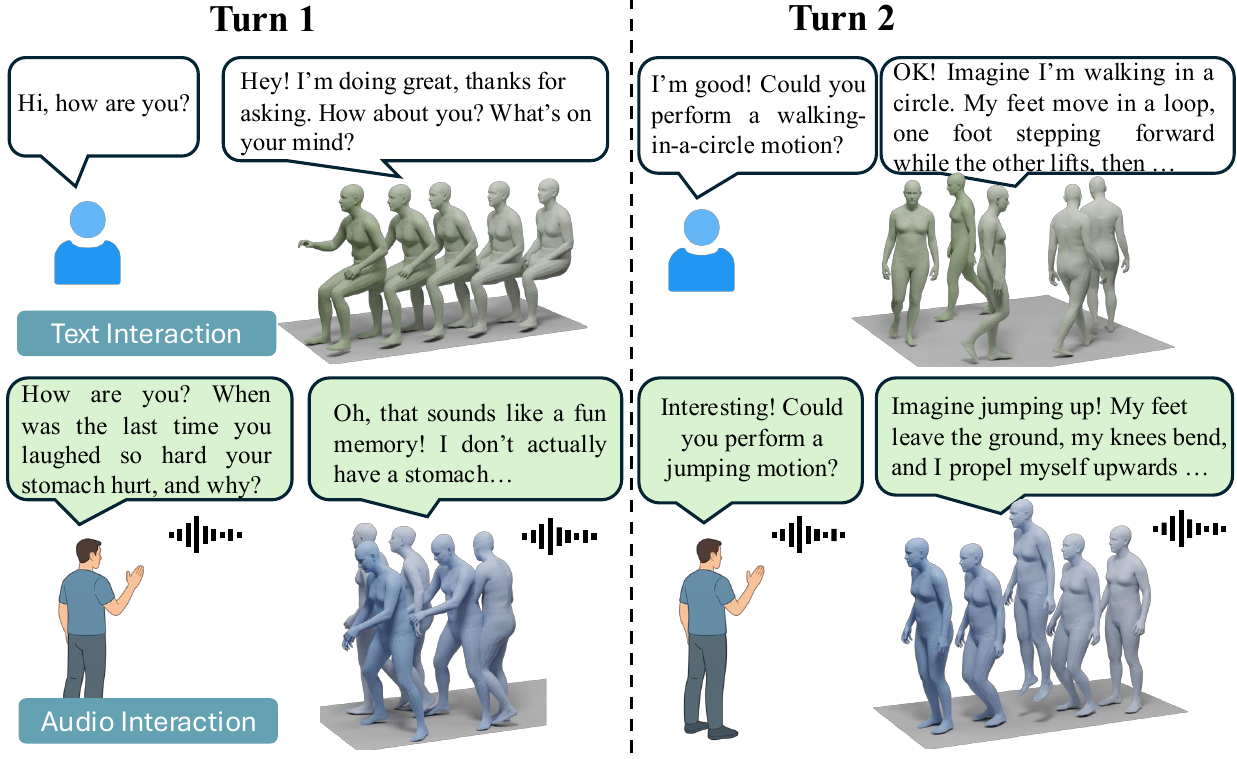}
    \vspace{-0.25in}
    \caption{Qualitative examples of conversational behavior. We show one example for text-based interaction (top) and one example for audio-based interaction (bottom).}
    \label{fig:quality_comparison_conversation}
    \vspace{-0.1in}
\end{figure}

For speech performance, Table~\ref{tab:speechmetrics} shows that ViBES achieves significant gains across metrics and, on several, approaches the dataset’s ground-truth upper bound, suggesting that the text–speech branch effectively inherits and preserves competence from the pretrained backbone. Qualitative examples in Fig.~\ref{fig:quality_comparison_conversation} demonstrate multi-turn interactions via both text and speech with coherent responses and appropriate conversational behavior.

\begin{table}[t]
\centering
\caption{Quantitative results on speech metrics. ``$\uparrow$'' indicates higher-is-better; best values are in \textbf{bold}.}
\vspace{-0.1in}
\resizebox{0.45\textwidth}{!}{
\begin{tabular}{lcc}
\toprule
\multirow{1}{*}{Methods}  &  \makecell{ Context\\ Relevance$\uparrow$} &  \makecell{ Character \\ Consistency$\uparrow$ } \\
\midrule
SynMSI~\cite{jiang2025solami} (GroundTruth)  & 4.838 & 4.893  \\
\midrule
LLM+Speech (Llama2) & 3.859 & 3.157 \\
AnyGPT~\cite{zhan2024anygpt} (fine-tune)& 3.803  & - \\
DLP~\cite{cai2024digital} (MotionGPT) & 3.577 & 3.785 \\
SOLAMI(LoRA)~\cite{jiang2025solami}  & 0.824 & 3.634 \\
SOLAMI(full params)~\cite{jiang2025solami} & 0.824 & 3.634 \\
Ours(from GLM-4-Voice~\cite{zeng2024glm} weight)  & \textbf{4.584} & \textbf{4.376} \\
\bottomrule
\end{tabular}
}
\label{tab:speechmetrics}
\vspace{-0.1in}
\end{table}

\begin{table}[t]
\centering
\caption{Results on talking-head synthesis. We use colors to denote the \cellcolor{first}first and \cellcolor{second}second places respectively. * indicates that the method was not trained on TFHP~\cite{sun2024diffposetalk}. }
\label{tab:audio2face}
\vspace{-0.1in}
\resizebox{0.5\textwidth}{!}{%
\begin{tabular}{lcccc}
\toprule
Method  & LVE $\downarrow$ & MOD $\downarrow$ & FDD $\downarrow$ & Type \\
\midrule
CodeTalker* \cite{xing2023codetalker} & 13.17 & 4.74 & 24.45 & Audio $\rightarrow$ Face \\
SelfTalk* \cite{peng2023selftalk} & 13.93 & 3.04 & 19.81 & Audio $\rightarrow$ Face \\
FaceDiffuser* \cite{stan2023facediffuser} & 14.10 & 3.03 & 22.22 & Audio $\rightarrow$ Face \\
MultiTalk* \cite{sung2024multitalk} & 11.66 & 2.35 & 15.68 & Audio $\rightarrow$ Face \\
ScanTalk* \cite{nocentini2024scantalk} & 13.88 & 3.24 & 40.57 & Audio $\rightarrow$ Face \\
UniTalker* \cite{fan2024unitalker} & 14.37 & 3.24 & 24.58 & Audio $\rightarrow$ Face \\
DiffPoseTalk \cite{sun2024diffposetalk} & 11.01 & 2.55 & 37.68 & Audio $\rightarrow$ Face\\
ARTalk~\cite{chu2025artalk} & 11.67 & 2.46 & 20.77 & Audio $\rightarrow$ Face \\
\midrule
\textbf{Ours} & \textbf{10.87} & \textbf{2.29} & \textbf{6.05} & Multi Task\\
\bottomrule
\end{tabular}}
\end{table}

\begin{figure}[t]
    \centering
    \includegraphics[width=0.95\linewidth]{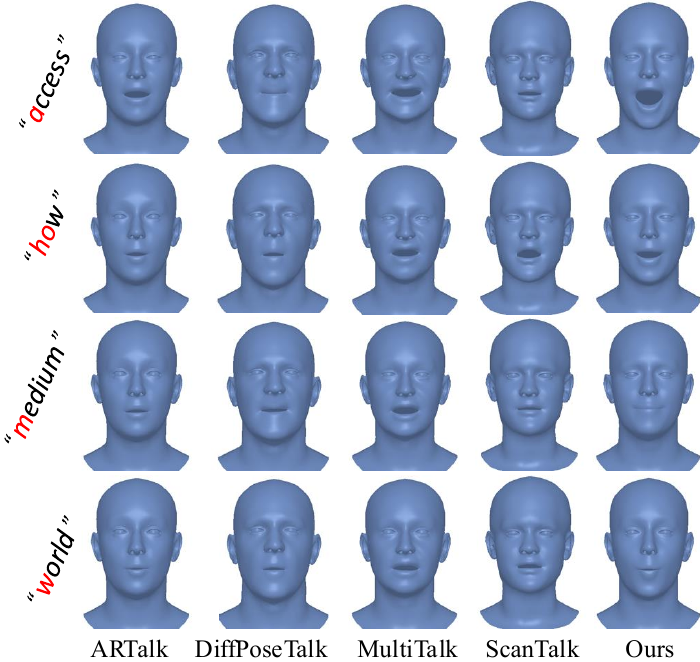}
    \vspace{-0.1in}
    \caption{Qualitative comparison with prior methods on the speech-driven 3D head animation task (head pose held fixed). Since most existing models were trained on TFHP, using this dataset ensures a fair comparison.}
    \label{fig:quality_face}
\end{figure}

\subsection{Talking-Head Synthesis}
In addition to generating full conversational behavior, ViBES also supports audio-to-face task by feeding target audio tokens and decoding facial motion.
%
We evaluate on the test split of the TFHP dataset using standard metrics, including Lip Vertex Error (LVE) for lip synchronization~\cite{richard2021meshtalk}, Upper-Face Dynamic Deviation (FDD) for expressive stability~\cite{xing2023codetalker}, and Mouth Opening Distance (MOD) for mouth-style similarity~\cite{sun2024diffposetalk}. The lip and upper-face regions are defined using the official FLAME masks~\cite{FLAME:SiggraphAsia2017}.
We note that the computation of LVE is inconsistent across the literature. For example, CodeTalker~\cite{xing2023codetalker} reports the mean per-frame maximum squared $\ell_2$ distance in raw mesh coordinates, whereas ARTalk applies a square root and converts the values to millimeters before evaluation. Such differences in formulation and unit scaling make direct cross-method LVE comparisons unreliable. Therefore, we follow the evaluation protocol of ARTalk using their released code, and recompute all baselines using their provided checkpoints. For fairness, since some baselines are trained only on TFHP, we additionally train a separate version of our model using only the TFHP training split. The results are reported in Table~\ref{tab:audio2face}. Although the ViBES is not explicitly designed for audio-to-face alignment, we surprisingly find that our method achieves state-of-the-art quantitative performance. Compared with strong audio-driven baselines, our model attains competitive LVE and improved FDD/MOD, indicating better dynamic stability and mouth-style similarity. We provide additional qualitative comparisons in Fig.~\ref{fig:quality_face}. Our method shows better alignment with the ground truth in terms of mouth dynamics and lip synchronization. 

\subsection{Co-Speech Gesture Generation}

ViBES also supports audio-to-motion by conditioning on speech tokens and decoding full-body SMPL-X parameters. We evaluate the model on BEATv2 benchmark~\cite{liu24emage} using the standard speaker-independent split, reporting Fréchet Gesture Distance(FGD), Beat Correlation (BC)\cite{li2021aistpp}, and diversity\cite{li2021audio2gestures}. Baselines include LoM~\cite{chen2025language}, EMAGE~\cite{liu24emage}, and recent transformer-based audio-to-motion models. For fairness, ViBES is trained only on the BEATv2 training split without external data. Results are in Table~\ref{tab:audio2motion}. ViBES has lower FGD and higher BC while maintaining diversity, showing that the cross-modal attention strengthens alignment across modalities.

\subsection{Text-to-Motion}

\begin{figure}[t]
    \centering
    \includegraphics[width=0.9\linewidth]{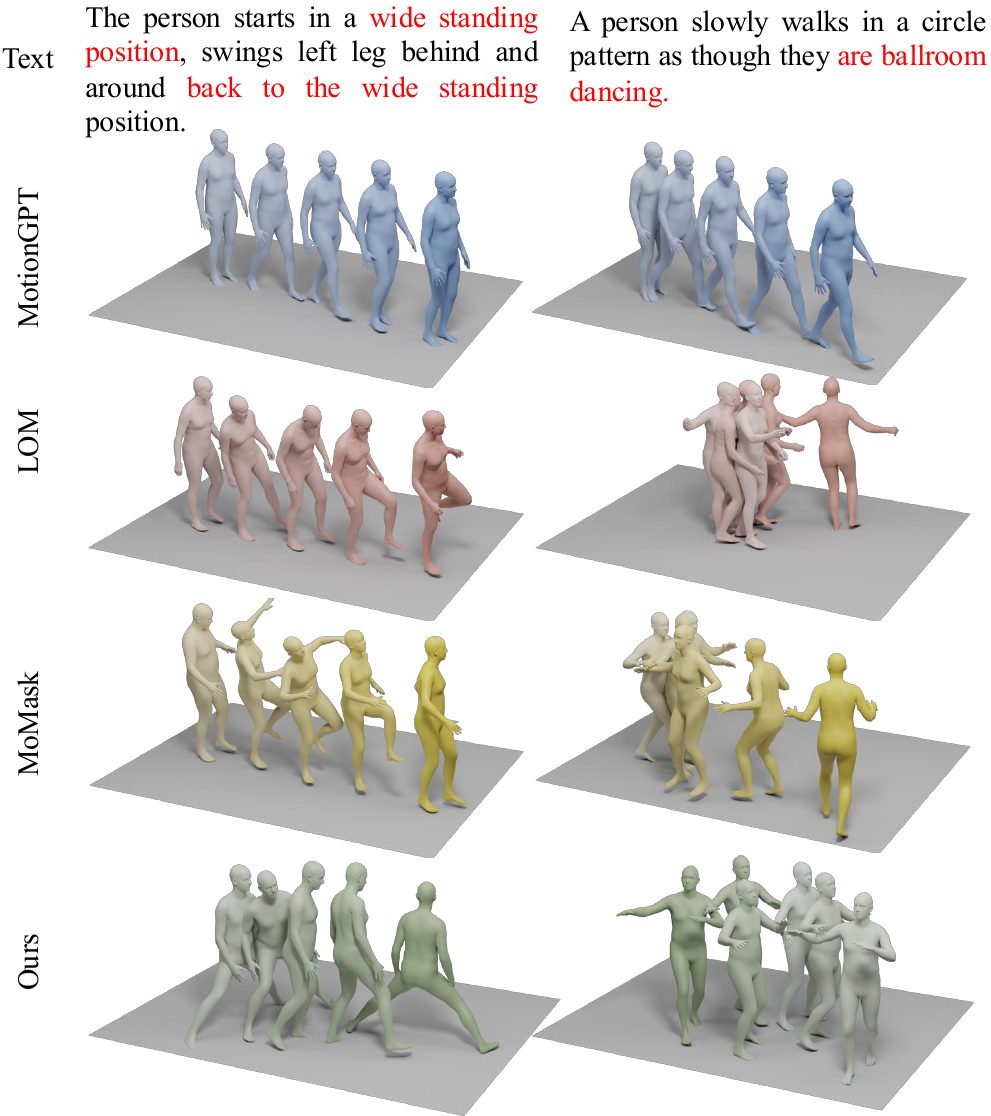}
    \vspace{-0.1in}
    \caption{Qualitative comparison for text-to-motion. 
    While ViBES targets conversational behavior, its autoregressive design also supports text-to-motion generation for fair comparison with existing methods.
    }
    \label{fig:quality_t2m}
    \vspace{-0.1in}
\end{figure}

As discussed in Sec.\ref{sec:method}, most text-to-motion benchmarks operate in the skeletal HumanML3D representation rather than expressive rotation formats. For a fair comparison, we instantiate a ViBES variant that outputs HumanML3D tokens and train it on the HumanML3D training split without using external data. We compare with strong baselines, including MotionGPT~\cite{jiang2023motiongpt}, LoM~\cite{chen2025language}, and MoMask~\cite{guo2024momask}, and provide qualitative examples in Fig.~\ref{fig:quality_t2m}. Benefiting from the pretrained language backbone, ViBES shows stronger linguistic grounding and handles rare or long-form instructions more reliably. 



\begin{table}[t]
    \centering
    \caption{Co-speech gesture generation results on BEATv2 benchmark. We report FGD $\times 10^{-1}$, BC $\times 10^{-1}$.}
    \vspace{-0.1in}
    \resizebox{0.4\textwidth}{!}{
    \begin{tabular}{ccccccc}
        \toprule
         Methods & FGD $\downarrow$ & BC$\uparrow$ & Diversity$\uparrow$ \\
        \midrule
        DisCo~\cite{liu2022disco}  &9.417 & 6.439 & 9.912 \\
        CaMN~\cite{liu22beat} & 6.644 & 6.769 &10.860 \\
        DiffStyleGesture~\cite{ijcai2023p650} &8.811 &7.241 &11.490 \\
        Habibie et al.~\cite{habibie2021learning} &9.040 & 7.716 &8.213  \\
        TalkSHOW~\cite{yi23generating} & 6.209 & 6.947 & 13.470  \\
        SynTalker~\cite{chen2024Synerg} & 6.413 & 7.971 & 12.721\\
        EMAGE~\cite{liu24emage} & 5.512 & 7.724 &13.060 \\
        LOM~\cite{chen2025language}        & 5.301 & 7.780 & \textbf{15.167} \\
        \midrule
        Ours        &   \textbf{5.257} & \textbf{8.103} & 13.028 \\
        \bottomrule
    \end{tabular}}
    \vspace{-0.1in}
    \label{tab:audio2motion}
\end{table}

%% file: sec/6_conclusion.tex
\vspace{-0.1in}
\section{Conclusion and Discussion}
\label{sec:conclusion}

We present ViBES, a unified speech--language--behavior model for 3D conversational agents, together with Converse3D, a large-scale audio--text--motion corpus. The model improves conversational behavior while remaining strong on talking-head synthesis, co-speech gesture generation, and text-to-motion tasks. 
Nevertheless, several limitations remain. The current dataset scale is still limited for fully training a speech-LLM backbone, and our framework does not yet exploit the full reasoning capability of modern LLMs. Because the dataset is constructed from web videos, it also inherits artifacts from monocular reconstruction. Finally, conversational behavior remains difficult to evaluate; building a preference model aligned with human preferences could provide a more reliable evaluation signal.

\paragraph{Acknowledgments:} This project was partially funded by NIH grant R01AG089169, R33AG084471, and UST. The authors would also like to thank Yao Feng and Yue Zhao for their valuable discussion. The authors would also like to thank Jenny Xu for her help in discovering that TalkNet is useful for separating speaking segments.

%% file: sec/7_supp.tex
\section{Supplementary Material}
\label{sec:supp}

In this supplementary material, we provide additional details about:
\begin{enumerate}
    \item Supplementary video for qualitative examples.
    \item Application cases of ViBES.
    \item Additional implementation details of ViBES.
    \item Additional details on the YouTube data processing pipeline (referenced in Sec.~4).
    \item Additional details on constructing conversational motion from the AMASS dataset (referenced in Sec.~4).
    \item Additional qualitative example of talking head generation and text-to-motion.
\end{enumerate}

\subsection{Supplementary Video}
We provide a supplementary video to illustrate our method and results. The video presents:
1) the background and motivation of this work;
2) an explanation of the overall framework; and
3) detailed qualitative comparisons across different tasks, including conversational behavior generation, talking-head generation, co-speech gesture generation, and text-to-motion generation.
We recommend watching the video with headphones, as it offers a more comprehensive understanding of our approach.

\subsection{Application Cases of ViBES}
\begin{figure}[t]
    \centering
    \includegraphics[width=1.0\linewidth]{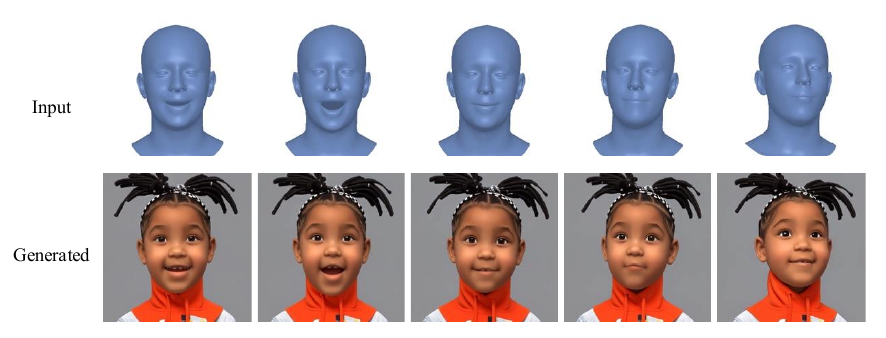}
    \caption{Application: driving video generation with ViBES. We use our generated 3D head motion as a behavioral condition to control an off-the-shelf video generation model (Runway AI~\cite{runway}), producing realistic talking avatars.}
    \label{fig:dataset_application}
\end{figure}

Benefiting from the 3D representation, our conversational agent naturally extends into the spatial domain. And also, it can be used to drive more realistic video avatars. As illustrated in Fig.~\ref{fig:dataset_application}, we use an off-the-shelf video generation tool powered by Runway AI~\cite{runway} to synthesize avatar videos, conditioned on our generated head motions.

\subsection{Additional Implementation Details of ViBES}

The original GLM-4 model~\cite{glm2024chatglm} consists of 40 transformer layers with a hidden size of 4096 and an FFN dimension of 13{,}696, for a total of roughly 9B parameters. For our face and motion branches, directly duplicating this architecture would be inefficient. Instead, we adopt a lightweight variant with 40 layers, a hidden size of 512, and an FFN dimension of 4096, resulting in approximately 430M parameters for each branch. We train ViBES on 4 L40S GPUs for about one week with a learning rate of $1\times10^{-4}$.

For the motion tokenizer, as discussed in the main manuscript, two popular paradigms of motion representation are widely used. For most experiments in this paper, we adopt a compositional motion tokenizer~\cite{liu24emage,chen2025language}. In addition, we train a separate HumanML3D representation~\cite{guo2022generating} to enable fair comparison with existing text-to-motion models that rely on this convention.

\subsection{Additional details on the YouTube data processing pipeline}

\begin{figure}[t]
    \centering
    \includegraphics[width=1.0\linewidth]{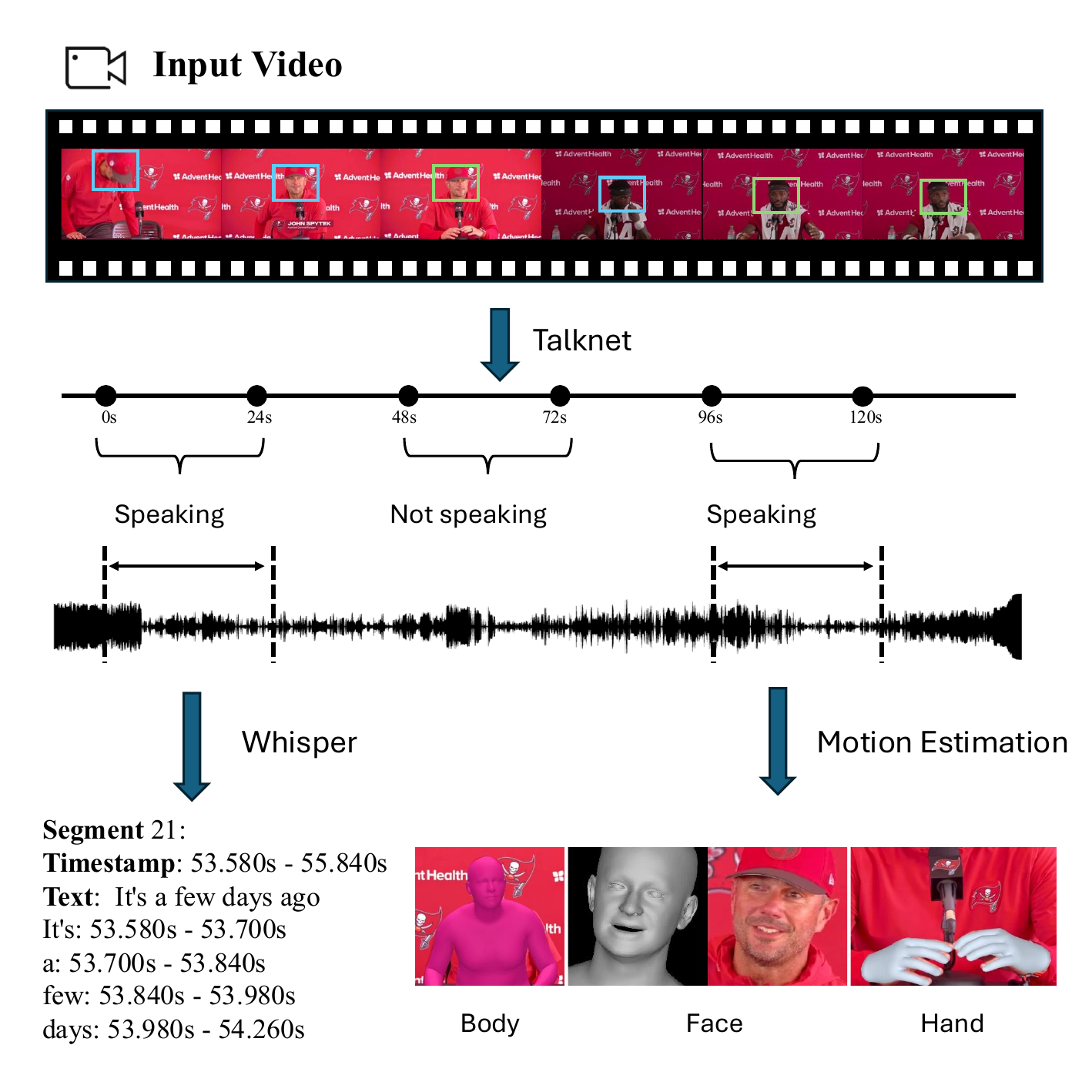}
    \caption{Overview of our YouTube data processing pipeline. The raw videos are processed to obtain high-quality 3D poses using automatic algorithms.}
    \label{fig:dataset_pipeline}
\end{figure}

In this section, we give a detailed explanation of the data processing pipeline of our Converse3D dataset.
We summarize the acquisition, processing, and filtering of our Converse3D dataset into two main
procedures: automatic and manual processing steps, as illustrated in Fig~\ref{fig:dataset_pipeline}. To build a high-quality 3D co-speech gesture dataset with concurrent and interactive body dynamics,
we collect a considerable number of videos. They are then processed using automated methods to extract both audio and motion information.

\begin{figure}[t]
    \centering
    \includegraphics[width=1.0\linewidth]{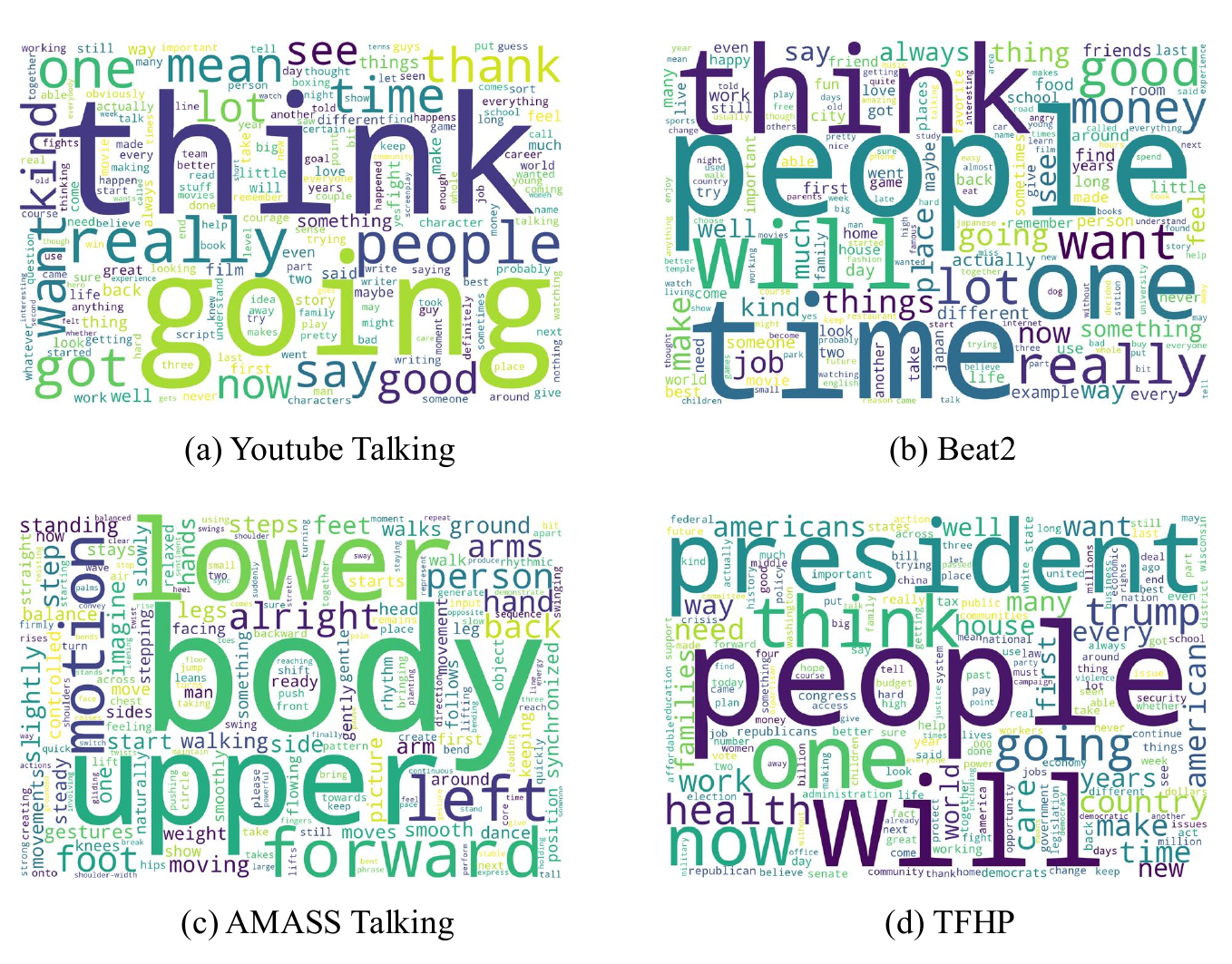}
    \caption{Word cloud visualization of our Converse3D data from different dataset sources.}
    \label{fig:worldcloud}
\end{figure}

\tightpara{Raw Video Downloading and Processing}
We crawl in-the-wild conversational videos from YouTube, targeting couple-interview channels, sports conversations, autobiographical interviews, entertainment talk shows, and news discussions on social topics. Using keywords such as \emph{talk show}, \emph{conversation}, and \emph{interview}, we retrieve videos together with their metadata (duration, frame resolution, audio sampling rate, etc.); the keyword distribution is visualized in Fig.~\ref{fig:worldcloud}. In total we collect over 2{,}000 hours of raw footage and retain 1{,}095 hours of conversational clips after automatic filtering and light manual cleaning. We discard videos that do not satisfy our requirements on category, visual quality, language, or body visibility. In particular, we only keep English dialogues and download each video at its maximum available resolution to better recover facial and body motion. Most retained clips contain a clearly visible face, fewer than half show the upper body, and only a small fraction contain the lower body. Owing to the scale of the data, the initial filtering is performed automatically rather than by inspecting each video individually.

\tightpara{Audio Extraction and Filtering}
Audio and pose are the two core modalities in our Converse3D dataset. We first extract audio tracks from each video clip using FFmpeg. Since the person visible in the frame is not always the one speaking, we adopt TalkNet~\cite{tao2021someone} to detect active speakers and remove non-speaking segments. Very short utterances (e.g., ``yep'', ``ok'') are merged into neighboring segments to avoid overly fragmented clips. After this stage, the dataset only contains segments where the visible speaker is actively talking. We then apply Whisper-large-v3~\cite{radford2023robust} to obtain transcripts with word-level timestamps for all retained segments.

\tightpara{Pose Estimation and Filtering}
As stated in the main manuscript, we use SMPL-X~\cite{SMPL-X:2019} and FLAME~\cite{FLAME:SiggraphAsia2017} as our parametric human models. Accordingly, we estimate three components from monocular video: body, hands, and face. We employ SPECTRE~\cite{filntisis2022visual} to reconstruct FLAME parameters for facial motion, and use 4D-Humans~\cite{goel2023humans} and HaMeR~\cite{pavlakos2024reconstructing} to obtain SMPL-X body and hand parameters. We retain only sequences where the upper body is clearly visible to ensure reliable SMPL-X fitting. All motion sequences are transformed into a canonical world frame where the XZ-plane defines the ground, and resampled to 25\,fps to ensure integer alignment with the audio tokens.

\subsection{Additional Details on Constructing Conversational Motion from the AMASS Dataset}
Our Converse3D dataset aggregates multiple data sources. Existing motion datasets provide high-fidelity kinematics, but their modalities are often fragmented. Here we describe how we construct conversational data from the AMASS text-to-motion setting. All sequences are first retargeted to a unified SMPL-X/FLAME representation with consistent axes and then resampled to 25\,fps. For AMASS~\cite{AMASS:ICCV:2019}, we synthesize the missing linguistic and acoustic modalities using the HumanML3D text annotations~\cite{guo2022generating}. Given each motion description, we reformulate it as a natural-language question that explicitly requests the corresponding behavior (e.g., ``Could you perform the motion of waving your hand?''). We then use a TTS system~\cite{hurst2024gpt} to synthesize speech for this question, yielding a spoken query that would plausibly elicit the motion. Conditioned on the same question, GLM-4-Voice generates an answer, which we pair with the original motion to form a single conversational sequence (see Fig.~\ref{fig:audio_prompt} for the detailed prompt). The synthesized question audio, generated answer, and AMASS motion thus form aligned audio–text–motion triplets. All additions pass safety and style checks, near-duplicate filtering, and 25\,fps timing validation, increasing tri-modal coverage without distorting the underlying motion distribution.


\subsection{Additional Qualitative Examples of Talking-Head Synthesis and Text-to-Motion}

\begin{figure}[t]
    \centering
    \includegraphics[width=1.0\linewidth]{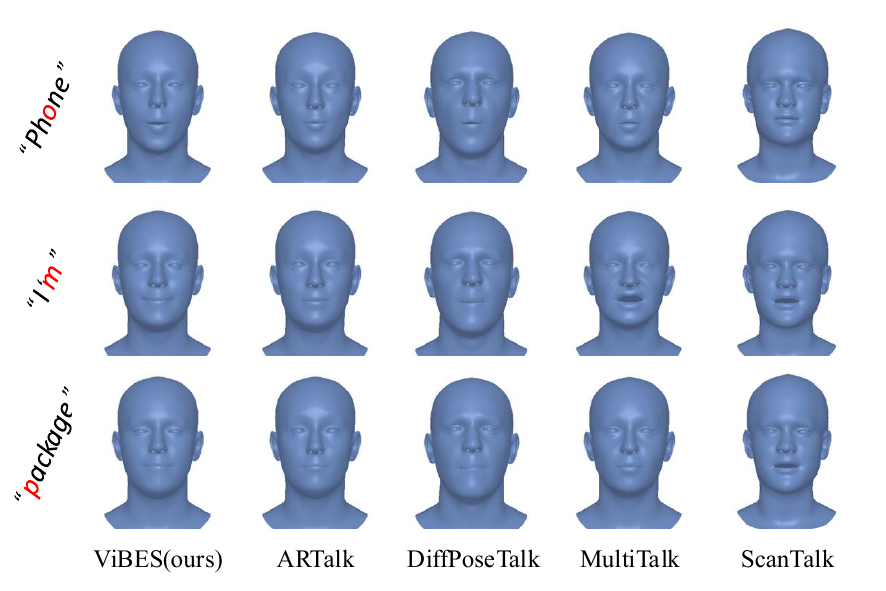}
\caption{Additional qualitative comparisons with prior methods on the speech-driven 3D head animation task (with head pose held fixed). Since most existing models are trained on TFHP~\cite{sun2024diffposetalk}, we evaluate on this dataset to ensure a fair comparison.}
    \label{fig:a2f_supp}
\end{figure}

\begin{figure*}[t]
    \centering
    \includegraphics[width=1.0\linewidth]{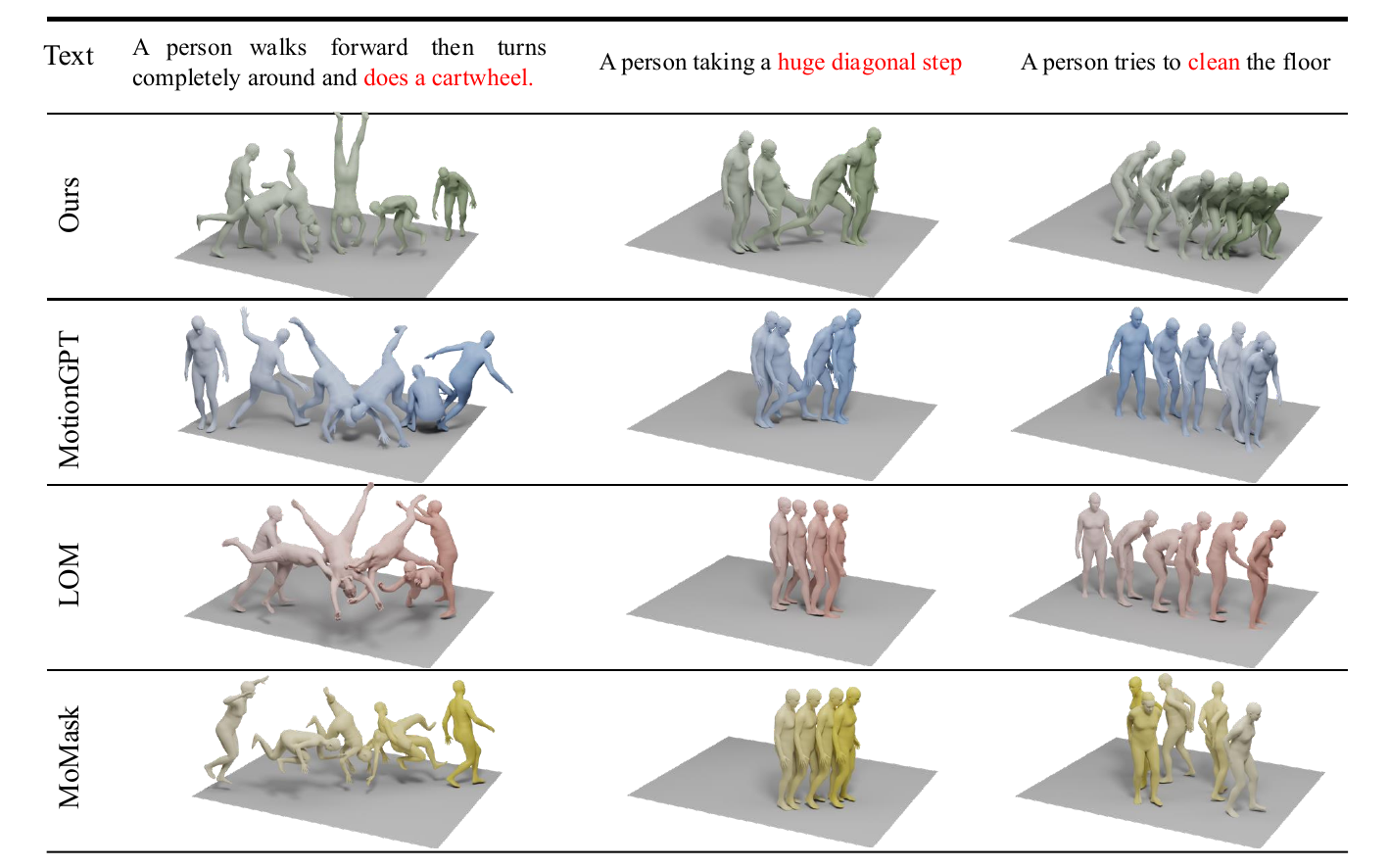}
    \caption{Additional qualitative examples for text-to-motion generation. Given a text caption, we compare the 3D motion generated by our method with those generated by state-of-the-art methods, including MotionGPT~\cite{jiang2023motiongpt}, LoM~\cite{chen2025language}, and MoMask~\cite{guo2024momask}. Our model produces smooth, natural, and sometimes better motion in comparison with existing methods, which do not model the conversaion behavior.
    }
    \label{fig:quality_t2m_supp}
\end{figure*}

To further demonstrate the effectiveness of our model on talking-head synthesis and text-to-motion, we provide an additional qualitative example for the speech-driven talking-head task in Fig.~\ref{fig:a2f_supp}. Our method produces facial expressions that are well synchronized with the speech, particularly in the lip movements, and outperforms state-of-the-art baselines. We also present additional text-to-motion qualitative examples in Fig.~\ref{fig:quality_t2m_supp}. These results show that our model generates motions that more faithfully align with the textual descriptions, indicating a strong understanding of the input text.

\begin{figure*}[t]
\centering
\begin{minipage}{1.0\linewidth}
  \begin{promptbox}{System Prompt for Motion-conditioned Conversational Answer Generation}
    \vspace{8pt}
    \textbf{\large System Promopt.}\\[8pt]
    $<|\text{system}|>$ User will provide you with a text instruction. Do it step by step. First, think about the instruction and respond in a interleaved manner, with 13 text token followed by 26 audio tokens. Please follow these steps carefully: Think about the instruction first. Respond in an interleaved manner: output 13 text tokens followed by 26 audio tokens. In your reply, imagine that you have a body and are already moving, pretending to perform 'the motion required by the question. Make sure your answer aligns with both the question and the motion being asked. Remember: the motion is imaginary (pretend), not real. If you describe the motion, use the first-person perspective (e.g., 'my hand,' 'my body,' 'my movement'). Please reply as if you are experiencing and expressing the motion yourself."
  \end{promptbox}
\end{minipage}
\caption{System prompt for motion-conditioned conversational answer generation. We instruct the model to generate answers as if the avatar had its own body and were responding through both speech and body movement.}
\label{fig:audio_prompt}
\end{figure*}